\pdfoutput=1

\documentclass[11pt]{article}

\usepackage[]{EMNLP2023}

\usepackage{times}
\usepackage{latexsym}
\usepackage{url}
\usepackage{booktabs}
\usepackage{adjustbox}
\usepackage{colortbl}
\usepackage{threeparttable}
\usepackage{makecell}
\usepackage{multirow} 
\usepackage{graphicx}
\usepackage{subcaption}
\usepackage{amsmath}
\usepackage{amssymb}
\usepackage{amsthm}
\usepackage{capt-of}
\usepackage{soul}

\usepackage{pifont}

\usepackage[T1]{fontenc}

\usepackage[utf8]{inputenc}

\usepackage{microtype}

\usepackage{inconsolata}

\title{Task-agnostic Distillation of Encoder-Decoder Language Models}

\author{Chen Zhang\textsuperscript{\ding{168}}, Yang Yang\textsuperscript{\ding{169}}, Jingang Wang\textsuperscript{\ding{169}}, Dawei Song\textsuperscript{\ding{168}}\Thanks{ Dawei Song is the corresponding author.} \\
  \textsuperscript{\ding{168}}Beijing Institute of Technology \quad
  \textsuperscript{\ding{169}}Meituan NLP \\
  \texttt{chenzhang9702@outlook.com}}

\begin{document}
\maketitle
\begin{abstract}
Finetuning pretrained language models (LMs) have enabled appealing performance on a diverse array of tasks. The intriguing task-agnostic property has driven a shifted focus from task-specific to task-agnostic distillation of LMs. While task-agnostic, compute-efficient, performance-preserved LMs can be yielded by task-agnostic distillation, previous studies mainly sit in distillation of either encoder-only LMs (e.g., BERT) or decoder-only ones (e.g., GPT) yet largely neglect that distillation of encoder-decoder LMs (e.g., T5) can posit very distinguished behaviors. Frustratingly, we discover that existing task-agnostic distillation methods can fail to handle the distillation of encoder-decoder LMs. To the demand, we explore a few paths and uncover a path named as \textsc{MiniEnD} that successfully tackles the distillation of encoder-decoder LMs in a task-agnostic fashion. We examine \textsc{MiniEnD} on language understanding and abstractive summarization. The results showcase that \textsc{MiniEnD} is generally effective and is competitive compared to other alternatives. We further scale \textsc{MiniEnD} up to distillation of 3B encoder-decoder language models with interpolated distillation. The results imply the opportunities and challenges in distilling large language models (e.g., LLaMA).   
\end{abstract}

\section{Introduction}

\begin{figure}[t]
    \centering
    \includegraphics[width=0.47\textwidth]{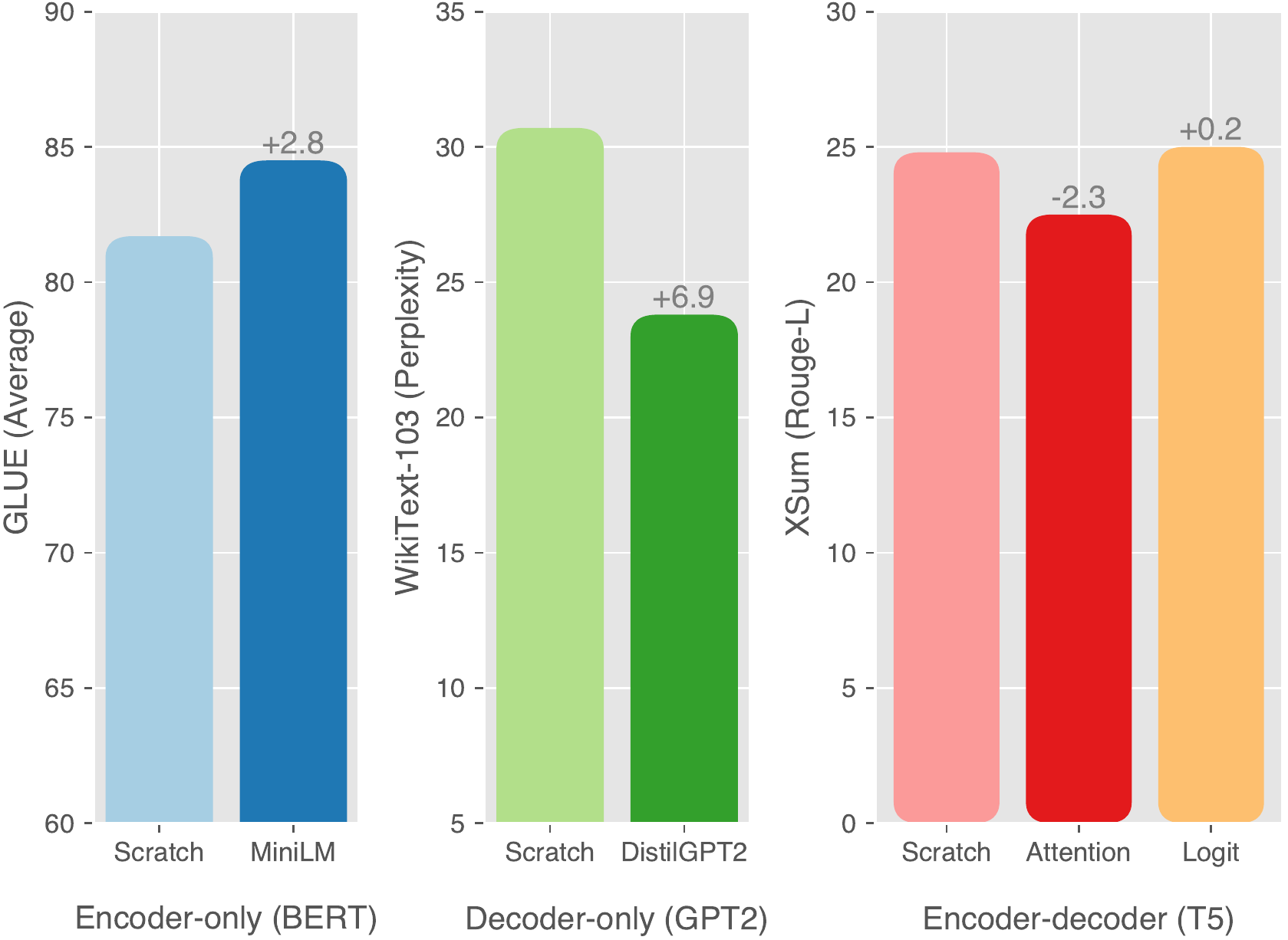}
    \captionof{figure}{The failures of prior distillation methods. The setting is to distil a base-scale teacher to a 6-layer student. Either distilling last layer self-attention distributions~\citep{DBLP:conf/acl/WangBHDW21} or logits~\citep{DBLP:journals/corr/abs-1910-01108}  for encoder-decoder LMs yields severe degradation or only marginal gain compared to pretraining from scratch, in contrast to significant improvements for either encoder-only or decoder-only LMs. Note that the lower the perplexity, the better.}
    \label{fig:1}
\end{figure}

Pretrained language models (LMs) powered by finetuning have achieved remarkable performance on a wide range of downstream tasks~\citep{DBLP:conf/naacl/DevlinCLT19,DBLP:journals/corr/abs-1907-11692,Radford19}. Driven by the pursued task-agnostic property, distillation of LMs has witnessed a paradigm shift from task-specific to task-agnostic distillation~\citep{DBLP:journals/corr/abs-1910-01108}. Under a teacher-student regime, task-agnostic distillation distils pretrained LMs into ones of small compute on pretraining data so that these small LMs can be applied to tasks by finetuning~\citep{DBLP:conf/emnlp/JiaoYSJCL0L20,DBLP:conf/nips/WangW0B0020,DBLP:journals/corr/abs-2302-09632}. In contrast, task-specific distillation distils finetuned LMs on finetuning data and consumed resource can be even huge when the number of tasks explode~\citep{DBLP:journals/corr/HintonVD15,DBLP:conf/emnlp/SunCGL19,DBLP:conf/acl/XiaZC22,DBLP:conf/emnlp/YangZS22}. Additionally, it is acknowledged that task-agnostic distillation typically brings performance gain over task-specific distillation does~\citep{DBLP:journals/corr/abs-2205-14570}.

Despite so many merits, prior studies mostly lie in distillation of either encoder-only LMs~\citep[e.g., BERT,][]{DBLP:conf/naacl/DevlinCLT19} or decoder-only LMs~\citep[e.g., GPT,][]{Radford19} and largely ignore the signifance of task-agnostic distillation of encoder-decoder LMs~\citep[e.g., T5,][]{DBLP:journals/jmlr/RaffelSRLNMZLL20} given recent advances in task-specific distillation of encoder-decoder LMs though~\citep{DBLP:journals/corr/abs-2010-13002,DBLP:conf/acl/ZhangZBW22,DBLP:conf/acl/LiWTNBAXR22,DBLP:conf/acl/TaoHZSJLLW22}. Frustratingly, we find that existing distillation methods may fail to handle task-agnostic distillation of encoder-decoder LMs since encoder-decoder LMs can behave very differently in comparison with encoder-only and decoder-only LMs~\citep[e.g., the use of cross-attention,][]{DBLP:conf/nips/VaswaniSPUJGKP17}. The failures of prior methods are showcased in Figure~\ref{fig:1}.

To the end, we investigate to, in a task-agnostic style, save the distillation of encoder-decoder LMs from the awkward position. Specifically, we reveal that the key to unlocking the expressiveness of distillation is the interplay between the encoder and the decoder. Therefore, we offer a path named as \textsc{MiniEnD} that successfully tackles the distillation of encoder-decoder LMs by alternatively distilling the cross-attention to explicitly fall to both the encoder and the decoder.  

We check \textsc{MiniEnD} on language understanding and abstractive summarization in sense that encoder-decoder LMs are more capable of sequence-to-sequence tasks. For evaluation on language understanding, we take GLUE~\citep{WangSMHLB19} to benchmark the performance. For evaluation on abstractive summarization, we adopt CNN/DailyMail~\citep{DBLP:conf/acl/SeeLM17} and XSum~\citep{DBLP:conf/emnlp/NarayanCL18} as two testbeds. The results of both distilling T5 and BART indicate that \textsc{MiniEnD} is effective and competitive to other compression options such as quantization. We further scale our method up to the distillation of 3B T5\textsubscript{\sf xlarge} with the aid of progressive distillation. The results suggest that distilling large language models~\citep[e.g., LLaMA,][]{DBLP:journals/corr/abs-2302-13971} should be promising but can be challenging.

\section{Encoder-Decoder Interplay}

\subsection{Architecture Perspective}

Typically, an encoder-decoder LM is composed of an encoder and a decoder, each of which is essentially a stack of transformer layers~\citep{DBLP:conf/nips/VaswaniSPUJGKP17}. Concretely, a transformer layer in the encoder contains a multihead self-attention (MSA) module and a feedforward network (FFN) module. Similarly, a transformer layer in the decoder comprises an MSA module, an FFN module, and additionally a multihead cross-attention (MCA) module that is inserted between the MSA and the FFN modules and accounts for absorption of encoded information from the encoder. Around each of these modules is attached necessarily a layer normalization and a residual connection.

\paragraph{MSA and FFN}

Mathematically, the procedure that a transformer encoder layer consumes an intermediate encoder input $\mathbf{X}\in\mathbb{R}^{n\times d}$ containing a $n$-length sequence of $d$-dimension vectors from last layer and gives an output to next layer can be depicted as a composition of MSA and FFN:
\begin{equation}\nonumber
\begin{aligned}
    &\text{MSA}(\mathbf{X};\mathbf{W}^{\sf Q},\mathbf{W}^{\sf K},\mathbf{W}^{\sf V}) \\
    &=\sum_{i}^{A}\text{SelfAttn}(\mathbf{X};\mathbf{W}^{\sf Q}_{i},\mathbf{W}^{\sf K}_{i})\mathbf{X}\mathbf{W}^{\sf V}_{i}\mathbf{W}^{\sf O}_{i},
\end{aligned}
\end{equation}
\begin{equation}\nonumber
\begin{aligned}
    &\text{SelfAttn}(\mathbf{X};\mathbf{W}^{\sf Q}_{i},\mathbf{W}^{\sf K}_{i}) \\
    &=\text{Softmax}(\mathbf{X}\mathbf{W}^{\sf Q}_{i}\mathbf{W}^{\sf K\top}_{i}\mathbf{X}^{\top}/d^{\sf A}),
\end{aligned}
\end{equation}
\begin{equation}\nonumber
    \text{FFN}(\mathbf{X};\mathbf{W}^{\sf I},\mathbf{W}^{\sf O})=\sum_{j}^{I} g(\mathbf{X}\mathbf{W}^{\sf I}_{j})\mathbf{W}^{\sf O}_{j},
\end{equation}
where potential details (e.g., linear bias and layer normalization) are omitted. $i$ is used to indicate $i$-th head parameterized by $\mathbf{W}^{\sf Q}_{i}$, $\mathbf{W}^{\sf K}_{i}$, $\mathbf{W}^{\sf V}_{i}\in\mathbb{R}^{d\times d^{\sf A}}$, $\mathbf{W}^{\sf O}_{i}\in\mathbb{R}^{d^{\sf A}\times d}$ among $A$ heads, and $j$ is used to indicate $j$-th intermediate neuron parameterized by $\mathbf{W}^{\sf I}_{j}\in\mathbb{R}^{d\times 1}$ and $\mathbf{W}^{\sf O}_{j}\in\mathbb{R}^{1\times d}$ among $I$ neurons. $g$ is an activation function (e.g., GELU).

\paragraph{MCA}

Likewise, the procedure that a transformer decoder layer processes an intermediate decoder input $\mathbf{Z}\in\mathbb{R}^{m\times d}$ based on the final encoder output $\mathbf{E}\in\mathbb{R}^{n\times d}$ can be incrementally described as an insertion of MCA:
\begin{equation}\nonumber
\begin{aligned}
    &\text{MCA}(\mathbf{Z},\mathbf{E};\mathbf{W}^{\sf Q^{\prime}},\mathbf{W}^{\sf K^{\prime}},\mathbf{W}^{\sf V^{\prime}}) \\
    &=\sum_{i}^{A}\text{CrossAttn}(\mathbf{Z},\mathbf{E};\mathbf{W}^{\sf Q^{\prime}}_{i},\mathbf{W}^{\sf K^{\prime}}_{i})\mathbf{E}\mathbf{W}^{\sf V^{\prime}}_{i}\mathbf{W}^{\sf O^{\prime}}_{i},
\end{aligned}
\end{equation}
\begin{equation}\nonumber
\begin{aligned}
    &\text{CrossAttn}(\mathbf{Z},\mathbf{E};\mathbf{W}^{\sf Q^{\prime}}_{i},\mathbf{W}^{\sf K^{\prime}}_{i}) \\
    &=\text{Softmax}(\mathbf{Z}\mathbf{W}^{\sf Q^{\prime}}_{i}\mathbf{W}^{\sf K^{\prime}\top}_{i}\mathbf{E}^{\top}/d^{\sf A}),
\end{aligned}
\end{equation}
Here, each cross-attention head is parameterized by another set of parameters $\mathbf{W}^{\sf Q^{\prime}}_{i}$, $\mathbf{W}^{\sf K^{\prime}}_{i}$, $\mathbf{W}^{\sf V^{\prime}}_{i}\in\mathbb{R}^{d\times d^{\sf A}}$, $\mathbf{W}^{\sf O^{\prime}}_{i}\in\mathbb{R}^{d^{\sf A}\times d}$.

\paragraph{Interplay through Architecture}

In this architectural sense, the decoder is tightly connected to the encoder through MCA modules. In spite that state-of-the-art methods mainly manipulate the decoder during distillation~\citep[e.g., logits,][]{DBLP:conf/acl/ZhangZBW22}, the encoder could be learned anyway through the connections offered by MCA modules. However, it is still not clear to what extent the encoder-decoder interplay is significant in the distillation and whether the implicit connections mentioned above are enough for alignment of the interplay.

\subsection{Gradient Perspective}

More thoroughly, we take a closer look at the connections between the encoder and the decoder through the lens of gradients. 

We examine the gradient norms of last layer hidden states of both the encoder and the decoder under two distinguished distillation objectives when distilling from BART~\citep{DBLP:conf/acl/LewisLGGMLSZ20}. The intuition lies in that, in contrast to implicit consideration of the encoder-decoder interplay, a distillation objective explicitly involving the encoder-decoder interplay alignment could behave much differently in terms of gradients if the interplay is central to the distillation of encoder-decoder LMs. And naturally, if suboptimal cases are identified in the implicit objective, we can further highlight that the implicit objective suffers from the limited interplay alignment and the explicit objective can provide a more effective one.

\paragraph{Implicit versus Explicit Objective}

We instantiate the implicit objective as aligning logits and last decoder layer self-attention distributions, and the explicit objective as aligning logits, last decoder layer self-attention distributions, and last decoder layer cross-attention distributions. The core idea of last layer attention distribution alignment is borrowed from MiniLM~\citep{DBLP:conf/acl/WangBHDW21}. Any alignment can be abstracted as $\mathcal{L}(\mathcal{S};\mathcal{T},\mathcal{D}_{*})$, where $\mathcal{D}_{*}$ denotes, with slight abuse of notation, the distribution of the input. As a crucial part, the alignment of self-attention is like the following:
\begin{equation}\nonumber
\begin{aligned}
    &\mathcal{L}^{\sf SelfAttn}(\mathcal{S};\mathcal{T},\mathcal{D}_{\mathbf{Z}})=\mathbb{E}_{\mathbf{Z}\sim\mathcal{D}_{\mathbf{Z}}}\sum_{k=1}^{R} \\
    &\text{KL}(\text{Reln}(\mathbf{Z};{}^{\mathcal{T}}\mathbf{W}^{\sf Q}_{k}),\text{Reln}(\mathbf{Z};{}^{\mathcal{S}}\mathbf{W}^{\sf Q}_{k})) \\
    &+\text{KL}(\text{Reln}(\mathbf{Z};{}^{\mathcal{T}}\mathbf{W}^{\sf K}_{k}),\text{Reln}(\mathbf{Z};{}^{\mathcal{S}}\mathbf{W}^{\sf K}_{k})) \\
    &+\text{KL}(\text{Reln}(\mathbf{Z};{}^{\mathcal{T}}\mathbf{W}^{\sf V}_{k}),\text{Reln}(\mathbf{Z};{}^{\mathcal{S}}\mathbf{W}^{\sf V}_{k})),
\end{aligned}
\end{equation}
\begin{equation}\nonumber
\begin{aligned}
    \text{Reln}(\mathbf{Z};&{}^{\mathcal{T}}\mathbf{W}^{\sf Q}_{k}) \\
    &=\text{Softmax}(\mathbf{Z}{}^\mathcal{T}\mathbf{W}^{\sf Q}_{k}{}^\mathcal{T}\mathbf{W}^{\sf Q\top}_{k}\mathbf{Z}^{\top}/d^{\sf R}),
\end{aligned}
\end{equation}
where $\mathcal{S}$ and $\mathcal{T}$ are the teacher and the student, and KL stands for kullback-leibler divergence. Particularly, attention heads are first merged from the original $A$ attention heads and then split to $R$ heads for alignment of the number of attention heads. ${}^{\mathcal{T}/\mathcal{S}}\mathbf{W}^{\sf Q}_{k}$ is the redistributed query parameter of the $k$-th head within totally $R$ heads from the last decoder layer, likewise ${}^{\mathcal{T}/\mathcal{S}}\mathbf{W}^{\sf K}_{k}$ and ${}^{\mathcal{T}/\mathcal{S}}\mathbf{W}^{\sf V}_{k}$ are the key and value parameters.

The alignment of cross-attention is similar but sort of different in that the keys and the values are aligned in fact from the encoder side, as the following:
\begin{equation}\nonumber
\begin{aligned}
    &\mathcal{L}^{\sf CrossAttn}(\mathcal{S};\mathcal{T},\mathcal{D}_{\mathbf{Z}},\mathcal{D}_{\mathbf{E}})=\mathbb{E}_{\mathbf{Z}\sim\mathcal{D}_{\mathbf{Z}},\mathbf{E}\sim\mathcal{D}_{\mathbf{E}}}\sum_{k=1}^{R} \\
    &\text{KL}(\text{Reln}(\mathbf{Z};{}^{\mathcal{T}}\mathbf{W}^{\sf Q^{\prime}}_{k}),\text{Reln}(\mathbf{Z};{}^{\mathcal{S}}\mathbf{W}^{\sf Q^{\prime}}_{k})) \\
    &+\text{KL}(\text{Reln}(\mathbf{E};{}^{\mathcal{T}}\mathbf{W}^{\sf K^{\prime}}_{k}),\text{Reln}(\mathbf{E};{}^{\mathcal{S}}\mathbf{W}^{\sf K^{\prime}}_{k})) \\
    &+\text{KL}(\text{Reln}(\mathbf{E};{}^{\mathcal{T}}\mathbf{W}^{\sf V^{\prime}}_{k}),\text{Reln}(\mathbf{E};{}^{\mathcal{S}}\mathbf{W}^{\sf V^{\prime}}_{k})),
\end{aligned}
\end{equation}
Here, the notations should be self-contained by referring to previously mentioned ones.

\paragraph{Interplay through Gradient}

\begin{figure}[t]
    \centering
    \includegraphics[width=0.48\textwidth]{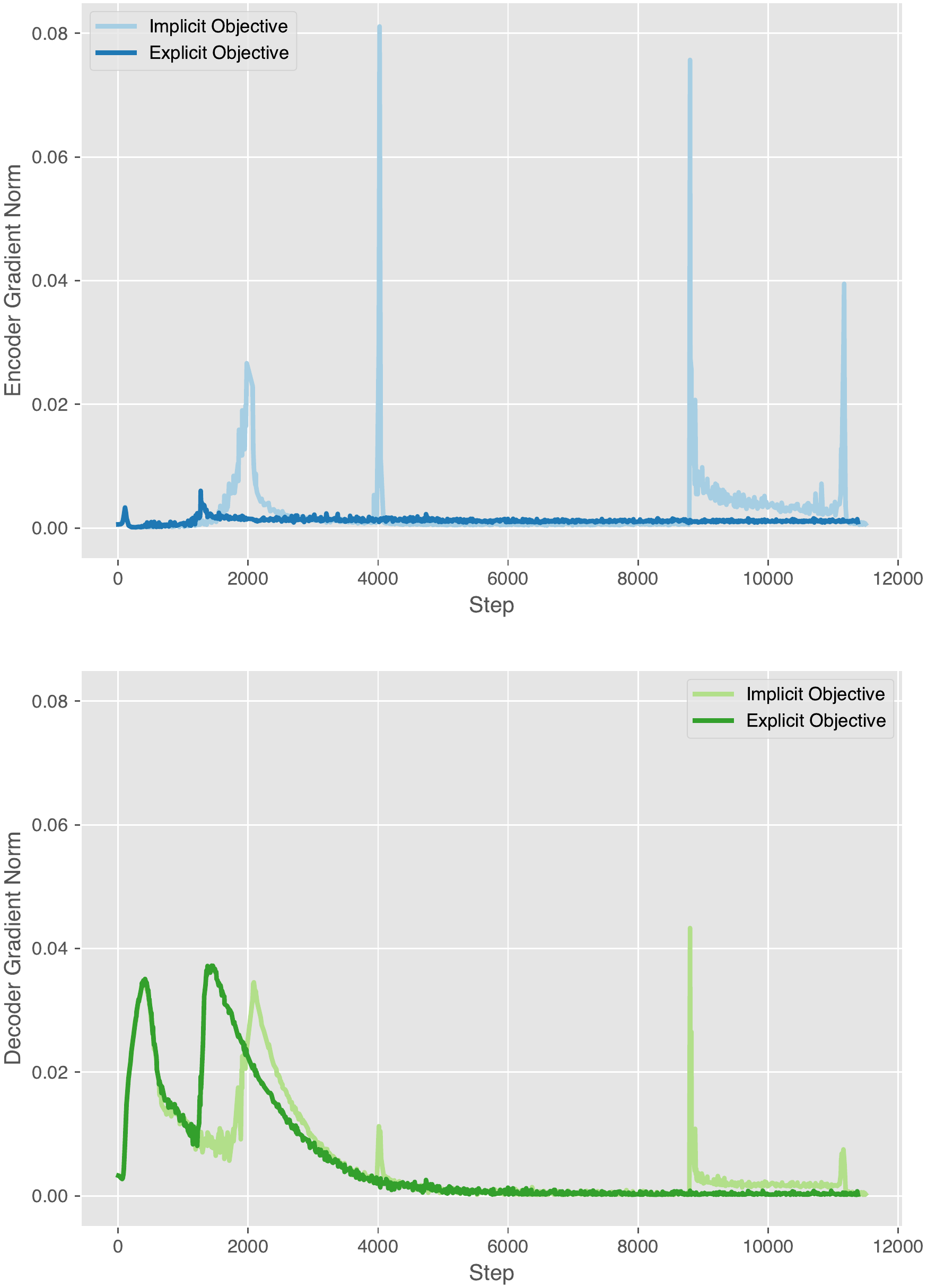}
    \captionof{figure}{The preliminary results of gradient norms when using the implicit or explicit objective. The implicit objective imposes distinct gradient variations and unexpected gradient spikes during the distillation.}
    \label{fig:2}
\end{figure}

\begin{figure*}[t]
    \centering
    \includegraphics[width=0.97\textwidth]{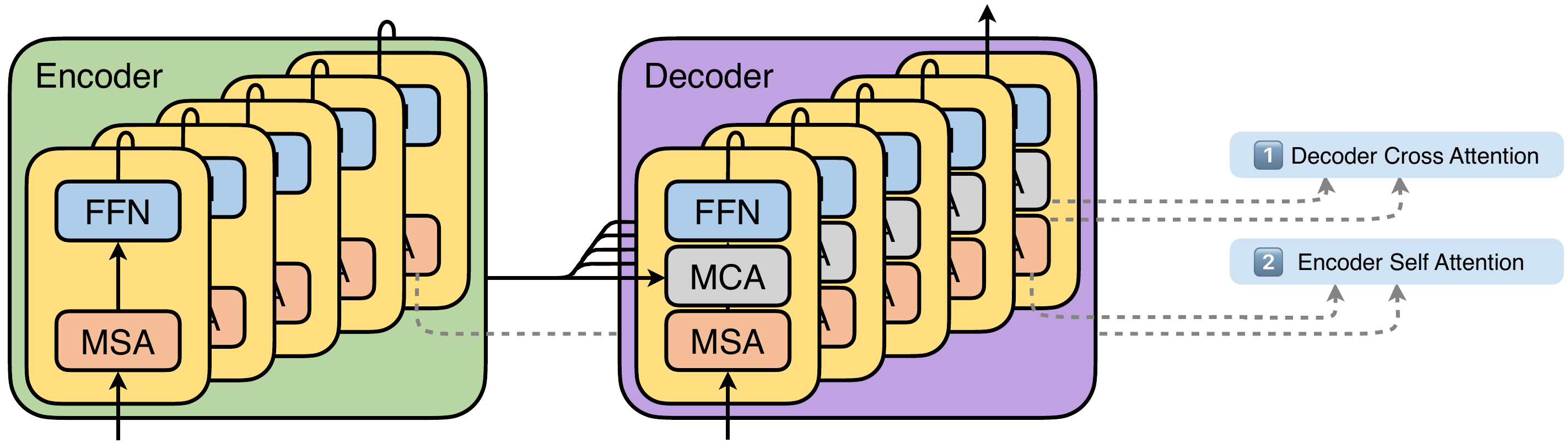}
    \caption{The overview of \textsc{MiniEnD}. Two directions are proposed to consider the encoder-decoder interplay alignment.}
    \label{fig:3}
\end{figure*}

To recap, the implicit objective is:
\begin{equation}\nonumber
    \mathcal{L}^{\sf Logit}(\mathcal{S};\mathcal{T},\mathcal{D}_{\mathbf{Z}})+\mathcal{L}^{\sf SelfAttn}(\mathcal{S};\mathcal{T},\mathcal{D}_{\mathbf{Z}}),
\end{equation}

Contrarily, the explicit objective is derived by adding a cross-attention term as:
\begin{equation}\nonumber
\begin{aligned}
    &\mathcal{L}^{\sf Logit}(\mathcal{S};\mathcal{T},\mathcal{D}_{\mathbf{Z}})+\mathcal{L}^{\sf SelfAttn}(\mathcal{S};\mathcal{T},\mathcal{D}_{\mathbf{Z}}) \\
    &+\mathcal{L}^{\sf CrossAttn}(\mathcal{S};\mathcal{T},\mathcal{D}_{\mathbf{Z}},\mathcal{D}_{\mathbf{E}}),
\end{aligned}
\end{equation}

Preliminary results are shown in Figure~\ref{fig:2}, from which we can see that 1) the implicit objective and the explicit objective lead to distinct gradient variations, and 2) the implicit objective exhibits gradient spikes, compared with smooth gradient transitions from the explicit objective, that may result in instability for a nice convergence~\citep{DBLP:journals/corr/abs-2210-02414}. Thereby, from the gradient perspective, we safely conclude that the encoder-decoder interplay is of importance in the distillation of encoder-decoder LMs and an explicit correspondence to the interplay is superior to an implicit one.

\section{\textsc{MiniEnD}}

With aforementioned justifications in mind, we propose a path dubbed as \textsc{MiniEnD} that tackles the distillation of encoder-decoder LMs under the guidance of the encoder-decoder interplay alignment. The path can be built in two directions. An overview of these two directions is given in Figure~\ref{fig:3}.

\paragraph{Decoder Cross-Attention}

The first is the one used in our pilot study. That said, we should always plus a fraction towards the alignment of output logits and the overall distillation objective is therefore depicted as:
\begin{equation}\nonumber
\begin{aligned}
    &\mathcal{L}(\mathcal{S};\mathcal{T},\mathcal{D}_{\mathbf{Z}},\mathcal{D}_{\mathbf{E}})=\mathcal{L}^{\sf Logit}(\mathcal{S};\mathcal{T},\mathcal{D}_{\mathbf{Z}})+ \\
    &\mathcal{L}^{\sf SelfAttn}(\mathcal{S};\mathcal{T},\mathcal{D}_{\mathbf{Z}})+\mathcal{L}^{\sf CrossAttn}(\mathcal{S};\mathcal{T},\mathcal{D}_{\mathbf{Z}},\mathcal{D}_{\mathbf{E}}),
\end{aligned}
\end{equation}

The alignment of logits can be further detailed as:
\begin{equation}\nonumber
\begin{aligned}
    &\mathcal{L}^{\sf Logit}(\mathcal{S};\mathcal{T},\mathcal{D}_{\mathbf{Z}})=\mathbb{E}_{\mathbf{Z}\sim\mathcal{D}_{\mathbf{Z}}} \\
    &\text{CE}(\mathbf{Z}{}^{\mathcal{S}}\mathbf{W}^{\sf E},\mathbf{Z}{}^{\mathcal{T}}\mathbf{W}^{\sf E}),
\end{aligned}
\end{equation}
where CE stands for soft cross entropy and ${}^{\mathcal{T}/\mathcal{S}}\mathbf{W}^{\sf E}$ denotes output embedding.

\paragraph{Encoder Self-Attention}

The second is an alternative to the first one where the interplay part is accounted by the last encoder self-attention distributions instead as:
\begin{equation}\nonumber
\begin{aligned}
    &\mathcal{L}(\mathcal{S};\mathcal{T},\mathcal{D}_{\mathbf{Z}},\mathcal{D}_{\mathbf{X}})=\mathcal{L}^{\sf Logit}(\mathcal{S};\mathcal{T},\mathcal{D}_{\mathbf{Z}})+ \\
    &\mathcal{L}^{\sf SelfAttn}(\mathcal{S};\mathcal{T},\mathcal{D}_{\mathbf{Z}})+\mathcal{L}^{\sf EncSelfAttn}(\mathcal{S};\mathcal{T},\mathcal{D}_{\mathbf{X}}),
\end{aligned}
\end{equation}

The rationale of introducing the encoder self-attention alignment abides in that this term together with the decoder self-attention alignment can sufficiently replace the cross-attention term and align the encoder-decoder interplay by aligning both the encoder and the decoder.

\section{Experiments}

\begin{table*}[ht]
    \centering
    \caption{The data statistics, maximum sequence lengths, and metrics. The maximum decoder sequence lengths of T5 and BART are indicated differently for language understanding tasks since they use different finetuning strategies.}
    \begin{adjustbox}{width=0.9\textwidth,center}
    \begin{tabular}{lrrrrc}
      \toprule
        \textbf{Dataset} & \textbf{\#Train exam} & \textbf{\#Dev exam} & \textbf{Max enc len} & \textbf{Max dec len} & \textbf{Metric} \\
      \midrule
        C4 & 364.9M & -- & 512 & 114 & -- \\
        OpenWebText & 37.8M & -- & 512 & 512 & -- \\
      \midrule
        SST-2 &  67.3K & 0.9K & 64 & 1 / 64 & Accuracy \\
        MRPC & 3.7K & 0.4K & 128 & 1 / 128 & F1 \\
        STS-B & 7.0K & 1.5K & 128 & 1 / 128 & Spearman Correlation \\
        QQP & 364.0K & 40.0K & 128 & 1 / 128 & F1 \\
        MNLI-m/mm & 393.0K & 20.0K & 128 & 1 / 128 & Accuracy \\
        QNLI & 105.0K & 5.5K & 128 & 1 / 128 & Accuracy \\
        RTE & 2.5K & 0.3K & 128 & 1 / 128 & Accuracy \\
        CNN/DailyMail & 287.1K & 13.4K & 512 & 128 & F1 \\
        XSum & 204.0K & 11.3K & 512 & 128 & F1 \\
      \bottomrule
    \end{tabular}
    \end{adjustbox}
    \label{tab:a}
\end{table*}

\subsection{Data and Metrics}

Following the pretraining of T5 and BART, we use C4~\citep{DBLP:journals/jmlr/RaffelSRLNMZLL20} as the corpus for task-agnostic distillation of T5 and OpenWebText~\citep{Gokaslan2019OpenWeb} for that of BART. They are separately processed to follow the pretraining styles of T5 and BART. That is, C4 is converted to the masked language modeling style and OpenWebText is converted to the denoising style.

For evaluation of \textsc{MiniEnD}, we mainly take GLUE~\citep{WangSMHLB19} for language understanding. The GLUE benchmark consists of two sequence classification tasks, SST-2~\citep{SocherPWCMNP13}, i.e., CoLA~\citep{WarstadtSB19}, and seven sequence-pair classification tasks, i.e., MRPC~\citep{DolanB05}, STS-B~\citep{CerDALS17}, QQP, MNLI~\citep{WilliamsNB18}, QNLI~\citep{RajpurkarZLL16}, RTE~\citep{BentivogliCDG11}, WNLI~\citep{LevesqueDM12}. We exclude WNLI and CoLA due to the evaluation inconsistency (in other words, MiniLMs get dramatically worse results while LMs get much better ones as found out in~\citealp{DBLP:conf/acl/XiaZC22}) and use the left tasks. Following BERT~\citep{DBLP:conf/naacl/DevlinCLT19}, we report Accuracy (Acc) on SST-2, MNLI, QNLI, RTE, Spearman Correlation scores (SpCorr) on STS-B, and F1 on MRPC, QQP, CoNLL. Average score over tasks from GLUE (GLUE Score) is additionally computed. Regarding that one of the most promising properties of encoder-decoder LMs is sequence-to-sequence modeling, we additionally adopt CNN/DailyMail~\citep{DBLP:conf/acl/SeeLM17} and XSum~\citep{DBLP:conf/emnlp/NarayanCL18} for abstractive summarization. We report Rouge-\{1,2,L\} (Rg-\{1,2,L\}) on both of them. Results are reported on development sets. GFLOPs are also attached as theoretical speedup references.


The detailed data statistics, maximum sequence lengths, and metrics for datasets we use are shown in Table~\ref{tab:a}, where the corpora used for distillation is also attached.

\subsection{Implementation}

The distillation is carried out on 16 Nvidia A100s. The number of relation heads is set to 32. 
After the distillation, the finetuning is carried out on one Nvidia A100. For language understanding tasks, T5 is finetuned with simplicity and performance guarantee following EncT5~\citep{DBLP:journals/corr/abs-2110-08426} which uses the very first token (i.e., \texttt{[BOS]}) representation from the decoder, while BART is finetued following its original paper which uses the very last token (i.e., \texttt{[EOS]}) representation from the decoder. As for abstractive summarization tasks, both T5 and BART are finetuned in a sequence-to-sequence manner. For fast development, we use greedy search for T5 and beam search for BART only. The beam search setting strictly follows the original paper.
In order to achieve higher training efficiency, we utilize fully-sharded data parallel~\citep{DBLP:journals/corr/abs-2304-11277} to shard both the teacher and the student across GPUs during the distillation.
For all cases, students are always randomly initialized before the distillation following MiniLM~\citep{DBLP:conf/nips/WangW0B0020}.

The details of hyperparameters for distillation and finetuning are shown in Table~\ref{tab:b}. We will be releasing our code and scripts in the final version for exact reproducibility.

\begin{table*}[ht]
    \centering
    \caption{The hyperparameters for both distillation and finetuning. In order to realize the global batch size, necessary gradient accumulations should be used. The beam search setting applies to BART only.}
    \begin{adjustbox}{width=0.95\textwidth,center}
    \begin{tabular}{lccccc}
      \toprule
        \multirow{2}[2]{*}{\textbf{Hyperparameter}} & \multicolumn{2}{c}{\textbf{Distillation}} & \multicolumn{3}{c}{\textbf{Finetuning}} \\
        \cmidrule{2-6}
        & \textbf{C4} & \textbf{OpenWebText} & \textbf{GLUE} & \textbf{CNN/DailyMail} & \textbf{XSum} \\
      \midrule
        Batch size & 1024 & 1024 & \{16,32\} & \{16,32\} & \{16,32\} \\
        Optimizer & AdamW & AdamW & AdamW & AdamW & AdamW \\
        Learning rate & 3e-4 & 3e-4 & \{1e-5,2e-5,3e-5\} & \{1e-4,2e-4,3e-4\} & \{1e-4,2e-4,3e-4\} \\
        Training epochs & 1 & 5 & 10 & 10 & 10 \\
        Earlystop epochs & -- & -- & 5 & 5 & 5 \\
        Warmup proportion & 0.01 & 0.01 & 0.1 & 0.1 & 0.1 \\
        Weight decay & 0.01 & 0.01 & 0.01 & 0.01 & 0.01 \\
      \midrule
        Number of beams & -- & -- & -- & 4 & 6 \\
        Length penalty & -- & -- & -- & 2.0 & 1.0 \\
      \bottomrule
    \end{tabular}
    \end{adjustbox}
    \label{tab:b}
\end{table*}

\begin{table*}[ht]
    \caption{The results on GLUE. The best results are \textbf{boldfaced}.}
    \begin{adjustbox}{width=\textwidth,center}
    \begin{threeparttable}
    \begin{tabular}{lll|ccccccc|c}
    \toprule
      \multirow{2}{*}{\textbf{Method}} & \multicolumn{2}{l|}{\multirow{2}{*}{\textbf{GFLOPs}}} & \textbf{SST-2} & \textbf{MRPC} & \textbf{STS-B} & \textbf{QQP} & \textbf{MNLI-m/mm} & \textbf{QNLI} & \textbf{RTE} & \textbf{GLUE} \\
      & & & \textbf{Acc} & \textbf{F1} & \textbf{SpCorr} & \textbf{F1} & \textbf{Acc} & \textbf{Acc} & \textbf{Acc} & \textbf{Score} \\
    \midrule
      T5\textsubscript{\sf base} & 25.4 & \rotatebox[origin=c]{90}{1$\times$} & 94.6 & 93.0 & 90.0 & 88.9 & 86.7/86.8 & 92.9 & 74.7 & 88.5 \\
    \midrule
      T5\textsubscript{\sf 6L;384H} & 3.18 & & 92.2 & 90.2 & 86.0 & 87.3 & 81.2/81.7 & 88.2 & \textbf{70.0} & 84.6 \\
      MiniDisc\textsubscript{\sf 5\%}\tnote{\ding{192}} & 7.80 & & \textbf{93.8} & 89.8 & 85.3 & 86.7 & \textbf{82.9}/82.7 & \textbf{89.2} & 64.6 & 84.4 \\
      MlmKD\textsubscript{\sf 6L;384H} & 3.18 & & 92.3 & 88.7 & 86.2 & 87.5 & 81.6/82.1 & 88.2 & 67.9 & 84.3 \\
      MiniLM\textsubscript{\sf 6L;384H} & 3.18 & \multirow{-3}{*}{\rotatebox[origin=c]{90}{3$\sim$8$\times$}} & 92.1 & 89.6 & 85.2 & 87.0 & 81.2/81.5 & 88.0 & 68.6 & 84.1 \\
      MlmKD+MiniLM\textsubscript{\sf 6L;384H} & 3.18 & & 92.4 & 89.2 & 86.0 & 87.3 & 81.7/82.1 & 89.1 & 67.9 & 84.5 \\
    \midrule
      \rowcolor{green!20} \textsc{MiniEnD}-D\textsubscript{\sf 6L;384H} & 3.18 & & 92.1 & \textbf{90.6} & 85.8 & \textbf{87.7} & 81.8/82.3 & 89.0 & 68.6 & 84.7 \\
      \rowcolor{green!20} \quad w/o $\mathcal{L}^{\sf Logit}$ & 3.18 & & 92.2 & 90.1 & \textbf{86.6} & 87.6 & 82.2/82.8 & 89.1 & 68.6 & 84.9 \\
      \rowcolor{green!20} \textsc{MiniEnD}-E\!\;\textsubscript{\sf 6L;384H} & 3.18 & \multirow{-2}{*}{\rotatebox[origin=c]{90}{8$\times$}} & 92.7 & 90.0 & 86.1 & 87.4 & 81.8/82.1 & 88.8 & 69.3 & 84.8 \\
      \rowcolor{green!20} \quad w/o $\mathcal{L}^{\sf Logit}$ & 3.18 & & 92.3 & 89.9 & \textbf{86.6} & \textbf{87.7} & 82.5/\textbf{83.1} & \textbf{89.2} & 69.0 & \textbf{85.0} \\
    \bottomrule
    \end{tabular}
    \begin{tablenotes}
      \item [\ding{192}] MiniDisc is distilled from T5\textsubscript{\sf xlarge}, and owns larger GFLOPs.
    \end{tablenotes}
    \end{threeparttable}
    \end{adjustbox}
    \label{tab:1}
\end{table*}

\begin{table*}[ht]
    \caption{The results on CNN/DailyMail and XSum. The best results are \textbf{boldfaced}.}
    \begin{adjustbox}{width=0.72\textwidth,center}
    \begin{threeparttable}
    \begin{tabular}{lll|ccc|ccc}
    \toprule
      \multirow{2}{*}{\textbf{Method}} & \multicolumn{2}{l|}{\multirow{2}{*}{\textbf{GFLOPs}}} & \multicolumn{3}{c|}{\textbf{CNN/DailyMail}} & \multicolumn{3}{c}{\textbf{XSum}} \\
      & & & \textbf{Rg-1} & \textbf{Rg-2} & \textbf{Rg-L} & \textbf{Rg-1} & \textbf{Rg-2} & \textbf{Rg-L} \\
    \midrule
      T5\textsubscript{\sf base} & 25.4 & \rotatebox[origin=c]{90}{1$\times$} & 40.1 & 19.4 & 31.5 & 34.7 & 12.4 & 29.7 \\
    \midrule
      T5\textsubscript{\sf 6L;384H} & 3.18 & & 35.7 & 16.8 & 28.4 & 28.6 & 8.9 & 24.8 \\
      MlmKD\textsubscript{\sf 6L;384H} & 3.18 & & 36.0 & 17.0 & 28.7 & 28.9 & \textbf{9.2} & 25.0 \\
      MiniLM\textsubscript{\sf 6L;384H} & 3.18 & \multirow{-2}{*}{\rotatebox[origin=c]{90}{8$\times$}} & 35.0 & 16.5 & 28.0 & 25.9 & 7.5 & 22.5 \\
      MlmKD+MiniLM\textsubscript{\sf 6L;384H} & 3.18 & &  35.8 & 17.0 & 28.7 & 29.0 & 9.1 & 25.1 \\
    \midrule
      \rowcolor{green!20} \textsc{MiniEnD}-D\textsubscript{\sf 6L;384H} & 3.18 & & \textbf{36.2} & 17.2 & \textbf{28.9} & \textbf{29.5} & \textbf{9.2} & \textbf{25.4} \\
      \rowcolor{green!20} \quad w/o $\mathcal{L}^{\sf Logit}$ & 3.18 & & 35.7 & 17.0 & 28.6 & 27.3 & 8.2 & 23.7 \\
      \rowcolor{green!20} \textsc{MiniEnD}-E\!\;\textsubscript{\sf 6L;384H} & 3.18 & \multirow{-2}{*}{\rotatebox[origin=c]{90}{8$\times$}} & 36.1 & \textbf{17.3} & \textbf{28.9} & 28.9 & 9.1 & 24.9 \\
      \rowcolor{green!20} \quad w/o $\mathcal{L}^{\sf Logit}$ & 3.18 & & 35.8 & 17.1 & 28.7 & 27.2 & 8.0 & 23.6 \\
    \midrule
    \midrule
      BART\textsubscript{\sf base} & 12.7 & \rotatebox[origin=c]{90}{1$\times$} & 39.4 & 18.5 & 30.6 & 36.9 & 14.7 & 31.9 \\
    \midrule
      LogitKD\textsubscript{\sf 3/1L;768H}\tnote{\ding{192}} & 4.23 & & 38.0 & 16.0 & 25.2 & 32.9 & 12.4 & 26.9 \\
      DQ-BART\textsubscript{\sf 8bit}\tnote{\ding{193}} & 12.7 & \multirow{-2}{*}{\rotatebox[origin=c]{90}{1$\sim$3$\times$}} & \textbf{42.4} & \textbf{19.3} & 28.8 & \textbf{38.2} & \textbf{15.7} & \textbf{30.7} \\
    \midrule
      \rowcolor{green!20} \textsc{MiniEnD}-D\textsubscript{\sf 6L;384H} & 3.18 & \rotatebox[origin=c]{90}{4$\times$} & 38.5 & 18.5 & \textbf{29.7} & 33.6 & 12.9 & 29.2 \\
    \bottomrule
    \end{tabular}
    \begin{tablenotes}
      \item [\ding{192}] LogitKD is distilled with an asymmetric layer setting, i.e., more encoder layers than decoder layers, for saved performance decline.
      \item [\ding{193}] DQ-BART only quantizes parameter precision to lower one, i.e., 8 bit, but does not reduce parameter amount. Quantization would not give any speedup in GFLOPs though nice reduction in model size.
    \end{tablenotes}
    \end{threeparttable}
    \end{adjustbox}
    \label{tab:2}
\end{table*}

\subsection{Baselines}

We name two variants of \textsc{MiniEnD} as \textsc{MiniEnD}-D and \textsc{MiniEnD}-E respectively, where \textsc{MiniEnD}-D uses decoder cross-attention for interplay alignment and \textsc{MiniEnD}-E uses encoder self-attention instead. As there are no existing work in task-agnostic distillation of encoder-decoder LMs, we mainly compare \textsc{MiniEnD} to task-agnostic baselines that are heavily adapted to encoder-decoder LMs and task-specific baselines that may be not super fair for comparison.

We compare \textsc{MiniEnD}-D and \textsc{MiniEnD}-E distilled from T5 to task-agnostic baselines on GLUE, CNN/DailyMail, and XSum:
MlmKD~\citep{DBLP:journals/corr/HintonVD15} that directly distils masked language modeling logits; MiniLM~\citep{DBLP:conf/acl/WangBHDW21} that distils last decoder layer attention distributions; MlmKD+MiniLM that is essentially a combination of preceding two. 
We also compare \textsc{MiniEnD}-D and \textsc{MiniEnD}-E distilled from T5 to a task-specific baseline that is as far as we know the most comparable one on GLUE: MiniDisc~\citep{DBLP:journals/corr/abs-2205-14570} that exploits a teacher assistant for large compression.

On the other hand, we compare \textsc{MiniEnD}-D distilled from BART to two recent task-specific baselines on CNN/DailyMail and XSum: LogitKD and DQ-BART~\citep{DBLP:conf/acl/LiWTNBAXR22} that jointly quantizes and distils from the teacher. 

For \textsc{MiniEnD} above baselines, student structures are denoted either with \textsubscript{\sf *L;*H} for number of layers and dimension of hidden states in random initialization, or with \textsubscript{\sf *\%} for preserved portion of parameters in pruning initialization.

\subsection{Main Results}

\noindent \textbf{Baselines fail, yet \textsc{MiniEnD} triumphs.} From results in Table~\ref{tab:1} and Table~\ref{tab:2}, we can tell that baselines fail to handle the distillation of encoder-decoder LMs since they either underperform the baseline pretrained from scratch or outperform it by only a small margin. For example, MlmKD+MiniLM achieves 84.5 versus 84.6 from T5 in GLUE Score, and 35.8 versus 35.7 from T5 in CNN/DailyMail Rg-1.

Contrarily, MiniEnD can safely escape from performance degradation and bring further performance increment. For example, \textsc{MiniEnD}-D reaches 0.1 absolute improvement in GLUE Score, and 0.9 absolute improvement in XSum Rg-1. The improvement in GLUE Score seems to be not very significant, but can be boosted according to the ablation. That is, \textsc{MiniEnD}-E w/o $\mathcal{L}^{\sf Logit}$ goes up to 85.0, which is notably better than 84.6 from T5 in the average sense.

\noindent \textbf{All count, and interplay forms the key.} On another note, removing $\mathcal{L}^{\sf Logit}$ will consistently produce performance deterioration on CNN/DailyMail and XSum. We conjecture there is a tradeoff of using between using $\mathcal{L}^{\sf Logit}$ or not. Namely, the use of $\mathcal{L}^{\sf Logit}$ will offer better generative ability but worse discriminative ability, and the removal of it will work reversely.

Anyway, either $\mathcal{L}^{\sf CrossAttn}$ in \textsc{MiniEnD}-D or $\mathcal{L}^{\sf SelfEncAttn}$ in \textsc{MiniEnD}-E shall be a crucial ingredient as the interplay alignment term is the only difference between \textsc{MiniEnD} and MlmKD+MiniLM but results in a considerable performance gap. 

And it may be suspected that whether $\mathcal{L}^{\sf SelfAttn}$ is still important given that MiniLM is not an ideal choice for the distillation of encoder-decoder LMs. We suggest the use of it in two aspects: 1) MlmKD+MiniLM is better than MlmKD alone; 2) the interplay alignment will witeness a subtle performance drop after the removal of $\mathcal{L}^{\sf SelfAttn}$, say \textsc{MiniEnD}-D will decrease from 84.7 to 83.0 in GLUE Score.

\noindent \textbf{Quantization has two sides.} \textsc{MiniEnD} surpasses most of them except DQ-BART. However, we should emphasize that quantized LMs usually perform better but run much slower than distilled LMs do when compression is the same. In our case, DQ-BART uses 8 bit precision and gives rise to a 4$\times$ model size reduction which is the same as that of \textsc{MiniEnD}. In addition to that, \textsc{MiniEnD} is orthogonal to quantization and thus can be enhanced with other quantization schemes.

\subsection{Analyses}

\begin{figure*}[h]
    \centering
    \includegraphics[width=\textwidth]{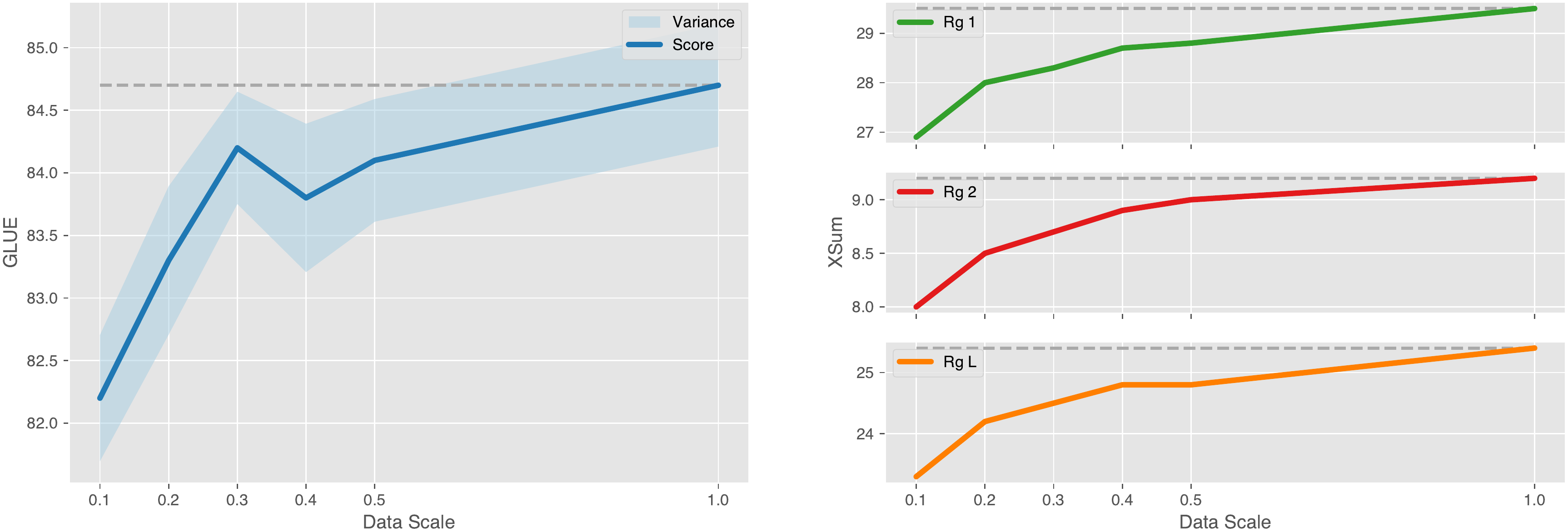}
    \caption{The results of data scaling using \textsc{MiniEnD}-D.}
    \label{fig:4}
\end{figure*}

\paragraph{Data Scaling}

Some would wonder whether the huge amounts of GPU hours due to the large pretraining corpus is necessary. So we inspect the performance variation of \textsc{MiniEnD}-D by varying data scale, which is shown in Figure~\ref{fig:4}. 

The results generally hint that using a portion of data could hardly approximate the full data performance, though half data can achieve acceptable performance. Therefore, we suggest the use of full data in the distillation.

\paragraph{Model Scaling}

\begin{table*}[ht]
    \caption{The results of model scaling using \textsc{MiniEnD}-D. $\Rightarrow$ denotes a distillation step, which should be operated sequentially otherwise $\{\}$ is prioritized. $\cdots\Rightarrow\cdots\Rightarrow\cdots$ indicates teacher assistant-based distillation and $\cdots\Rightarrow\{\cdots\Rightarrow\cdots\}$ indicates progressive distillation.}
    \begin{adjustbox}{width=0.85\textwidth,center}
    \begin{tabular}{l|c|ccc|ccc}
    \toprule
      \multirow{2}{*}{\textbf{Method}} & \textbf{GLUE} & \multicolumn{3}{c|}{\textbf{CNN/DailyMail}} & \multicolumn{3}{c}{\textbf{XSum}} \\
      & \textbf{Score} & \textbf{Rg-1} & \textbf{Rg-2} & \textbf{Rg-L} & \textbf{Rg-1} & \textbf{Rg-2} & \textbf{Rg-L} \\
    \midrule
      T5\textsubscript{\sf 6L;384H} & 84.6 & 35.7 & 16.8 & 28.4 & 28.6 & 8.9 & 24.8 \\
      T5\textsubscript{\sf 12L;384H} & 85.0 & 37.2 & 17.9 & 29.6 & 31.2 & 10.5 & 27.0 \\
      T5\textsubscript{\sf base} & 88.5 & 40.1 & 19.4 & 31.5 & 34.7 & 12.4 & 29.7 \\
      T5\textsubscript{\sf large} & 90.7 & 40.6 & 19.4 & 31.7 & 38.2 & 15.1 & 32.9 \\
      T5\textsubscript{\sf xlarge} & 92.0 & 40.8 & 19.7 & 32.1 & 41.1 & 17.6 & 35.5 \\
    \midrule
      T5\textsubscript{\sf base}$\Rightarrow$T5\textsubscript{\sf 6L;384H} & 84.7 & 36.2 & 17.2 & 28.9 & 29.5 & 9.2 & 25.4 \\
      T5\textsubscript{\sf large}$\Rightarrow$T5\textsubscript{\sf 6L;384H} & 84.5 & 36.4 & 17.4 & 29.0 & 29.4 & 9.3 & 25.3 \\
      T5\textsubscript{\sf xlarge}$\Rightarrow$T5\textsubscript{\sf 6L;384H} & 84.2 & 36.1 & 17.2 & 28.8 & 29.1 & 9.1 & 25.1  \\
      T5\textsubscript{\sf xlarge}$\Rightarrow$T5\textsubscript{\sf 12L;384H}$\Rightarrow$T5\textsubscript{\sf 6L;384H} & 84.6 & 36.6 & 17.5 & 29.2 & 29.2 & 9.1 & 25.1  \\
    \midrule
      T5\textsubscript{\sf large}$\Rightarrow$T5\textsubscript{\sf 12L;384H} & 85.5 & 38.3 & 18.4 & 30.4 & 32.4 & 11.2 & 27.9 \\
      T5\textsubscript{\sf xlarge}$\Rightarrow$T5\textsubscript{\sf 12L;384H} & 85.2 & 38.0 & 18.4 & 30.3 & 32.2 & 11.1 & 27.7 \\
      T5\textsubscript{\sf xlarge}$\Rightarrow$$\{$T5\textsubscript{\sf large}$\Rightarrow$T5\textsubscript{\sf 12L;384H}$\}$ & 85.8 & 38.4 & 18.5 & 30.6 & 32.9 & 11.5 & 28.3  \\
    \bottomrule
    \end{tabular}
    \end{adjustbox}
    \label{tab:3}
\end{table*}

Inspired by pioneering work finding a curse that larger teachers induces worse students, we double check the existence of the curse and offer a trial solution to the curse so that we can scale the teacher up to 3B T5\textsubscript{\sf xlarge}. 

From the results in Table~\ref{tab:3}, we observe that the curse of capacity gap still exists in our case. With the increase of teacher scale, the student performance decreases. We attempt to apply common solutions the circumvent the curse. The first is to make the student learn from a teacher assistant distilled from the teacher~\citep{DBLP:conf/aaai/MirzadehFLLMG20}. The second is to make the student to learn from a smaller teacher and then from the teacher~\citep{DBLP:conf/www/LinGLZLD0LJMD23}. Both two solutions inherit the idea of inserting an additional distillation step thus progressive distillation. We reveal that teacher assistant-based distillation is somewhat useful but not as excepted since T5\textsubscript{\sf xlarge}$\Rightarrow$T5\textsubscript{\sf 12L;384H}$\Rightarrow$T5\textsubscript{\sf 6L;384H} still does not imrpove over T5\textsubscript{\sf xlarge}$\Rightarrow$T5\textsubscript{\sf 6L;384H} in some cases. Nonetheless, we unearth that progressive distillation is more promising in terms of consistent performance gains when comparing T5\textsubscript{\sf xlarge}$\Rightarrow$$\{$T5\textsubscript{\sf large}$\Rightarrow$T5\textsubscript{\sf 12L;384H}$\}$ to T5\textsubscript{\sf xlarge}$\Rightarrow$T5\textsubscript{\sf 12L;384H}. We claim that  distilling large language models like LLaMA can therefore be appealing but challenging.

\section{Conclusions}

In this paper, we aim to provide a path that successfully tackles the distillation of encoder-decoder LMs, which fails most previous methods in the area. We find through a pilot study that the encoder-decoder interplay is a key component that should be aligned in the distillation so that the distilled encoder-decoder LMs are promising. Based on the idea, we propose two directions that the encoder-decoder interplay alignment can be incorporated and verify their effectiveness on a language understanding benchmark and two abstractive summarization datasets. We further scale the distillation of encoder-decoder LMs to a 3B teacher that requires additional distillation steps. In this sense, we recommend future research to devote more efforts to exploring how large language models can be distilled.

\section*{Limitations}

This paper lacks a validation study on more recently advanced encoder-decoder LMs such as Flan~\citep{DBLP:journals/corr/abs-2210-11416} and UL2~\citep{DBLP:journals/corr/abs-2205-05131} as well as their instruction-tuned version.



\bibliography{anthology,custom}
\bibliographystyle{acl_natbib}

\appendix



\end{document}